\documentclass[sigconf]{acmart}

\usepackage{booktabs} 
\usepackage{diagbox}
\usepackage{subfig}
\usepackage{graphicx}
\usepackage{caption}
\usepackage{hhline}
\usepackage{lipsum}
\usepackage{algorithm}
\usepackage[noend]{algpseudocode}

\setcopyright{acmcopyright}

\copyrightyear{2018} 
\acmYear{2018} 
\acmConference[KDD 2018]{24th ACM SIGKDD International Conference on Knowledge Discovery \& Data Mining}{August 19--23, 2018}{London, United Kingdom}
\acmBooktitle{KDD 2018: 24th ACM SIGKDD International Conference on Knowledge Discovery \& Data Mining, August 19--23, 2018, London, United Kingdom}
\acmPrice{15.00}
\acmDOI{10.1145/3219819.3219902} 
\acmISBN{978-1-4503-5552-0/18/08} 

\begin{document}
\title{Exploring Student Check-In Behavior for Improved Point-of-Interest Prediction}

\author{Mengyue Hang} 
\affiliation{%
  \institution{Computer Science Department}
  \city{Purdue University}
}
\email{hangm@purdue.edu}

\author{Ian Pytlarz}
\affiliation{%
  \institution{Office of Institutional Research}
  \city{Purdue University}
}
\email{ipytlarz@purdue.edu}

\author{Jennifer Neville}
\affiliation{%
  \institution{Computer Science and Statistics Departments}
  \city{Purdue University}
}
\email{neville@purdue.edu}

\renewcommand{\shortauthors}{M. Hang et al.}

\begin{abstract}
With the availability of vast amounts of user visitation history on \emph{location-based social networks} (LBSN), the problem of {\em Point-of-Interest} (POI) prediction has been extensively studied. However, much of the research has been conducted solely on voluntary check-in datasets collected from social apps such as Foursquare or Yelp. While these data contain rich information about recreational activities (e.g., restaurants, nightlife, and entertainment), information about more prosaic aspects of people's lives is  sparse. This not only limits our understanding of users' daily routines, but more importantly the modeling assumptions developed based on characteristics of recreation-based data may not be suitable for richer check-in data. 
In this work, we present an analysis of education ``check-in'' data using WiFi access logs collected at Purdue University. 
We propose a heterogeneous graph-based method to encode the correlations between users, POIs, and activities, and then jointly learn embeddings for the vertices. We evaluate our method compared to previous state-of-the-art POI prediction methods, and show that the assumptions made by previous methods significantly degrade performance on our data with dense(r) activity signals. We also show how our learned embeddings could be used to identify similar students (e.g., for friend suggestions).
\end{abstract}

%
%


\keywords{Location-based social networks, network embedding, heterogeneous graphs, representation learning.}

\maketitle

\newcommand\mynote[1]{\textcolor{blue}{#1}}
\newcommand{\update}[1]{[{\color{red}#1}]}
\newcommand{\model}{EDHG}

\section{Introduction}
Millions of check-in records in \emph{location-based social networks} (LBSNs) provide an opportunity to study users' mobility pattern and social behavior from a spatial-temporal perspective. In recent years, the \emph{point-of-interest} (POI) recommendation/prediction problem has attracted significant attention \cite{yang2017bridging}, \cite{yao2018exploiting}, \cite{he2016inferring}, \cite{yin2017spatial}, \cite{wang2016spore}, particularly for advertising and personalization.
In POI tasks, the goal is to use user behavioral data to model users' activities at different locations and times, and then make predictions (or recommendations) for relevant venues based on their current context (including spatial, temporal, and other contextual information). 

While POI predictions have broad applicability to myriad organizations, to date research has focused on developing POI methods based solely on voluntary check-in datasets collected from online social network apps such as Foursquare or Yelp \cite{noulas2011empirical}, \cite{kylasa2016social}. While these data contain rich information about recreational activities (e.g., restaurants, nightlife, and entertainment), the reliance on voluntary reporting results in sparse information about more prosaic aspects of daily life (e.g., offices, errands, houses). 
%
Moreover, recreation-based check-in data may bias conclusions drawn about mobility patterns or personal preferences. For example, Foursquare users often visit a POI only once, so the users' check-ins may not be sufficient to derive preferences for venues themselves, but only for venue {\em categories}. Also since check-ins to location-based social networks are often sporadic \cite{noulas2011empirical}, it can be difficult to identify consistent user patterns. 

In this work, we present the first analysis of a spatio-temporal educational ``check-in'' dataset, with the aim of using POI predictions to personalize student recommendations (e.g., clubs, friends, study locations) and to understand behavior patterns that increase student retention and satisfaction. The results also provide a better idea of how campus facilities are utilized and how students connect with each other. 
The Purdue University ``check-in'' data 
records (anonymized) users' access to WiFi access points on campus, with venue information about locations (e.g., dining hall, library, dorm, gym). Specifically, we analyze WiFi access history across on-campus buildings, for all freshmen over one semester. 

Compared to well-known check-in datasets like Foursquare, these data contain (1) more active users, (2) a richer set of daily activities (e.g., study, dine, exercise, rest), and (3) well-annotated spatial range (i.e., on campus). These characteristics make it easier to analyze the unique properties of user check-in data and extract interesting social and mobility patterns. 
Notably the WiFi access logs provide better temporal resolution than previous LBSN datasets, since a user ``checks-in'' whenever her device sends or receives a packet through a wireless connection. Similar data are collected by GPS trackers, where location observations are passively recorded~\cite{zheng2009mining}. But while GPS tracking provides more extensive information about users' movements, it does not provide the rich venue and activity information associated with check-in data. 


POI prediction and recommendation tasks are different from more traditional recommendation tasks because they involve a more structured, context-rich environment~\cite{sun2017recommendation}.
In addition to user-POI check-in frequencies, the users and POIs are usually associated with a rich set of attributes, such as POI category, spatio-temporal information, personal activity. A heterogeneous graph structure is thus a natural choice to for spatio-temporal POI prediction tasks, since it is more amenable to representing and reasoning with rich context compared to tensor factorization methods (e.g., \cite{he2016inferring}). 

Recently, methods which learn graph representations by {\em embedding} nodes in a vector space have gained traction from the research community, and graph embedding methods have been widely adopted for a variety of tasks, including text mining \cite{tang2015pte}, online event detection \cite{zhang2017react} and author identification \cite{chen2017task}.
In this work, we extend these efforts and propose a network-based embedding method called {\em Embedding for Dense Heterogeneous Graphs} (\model). Our approach (i) incorporates personal preferences, temporal patterns, and activity types into a sparse(r) view of the heterogeneous graph, (ii) uses global knowledge of the graph to generate negative samples, (iii) jointly learns vector representations for the nodes in the graph, i.e., users, POIs, time-slots, and then (iv) uses the learned representations for user and time specific POI recommendation. 

We empirically evaluate the effectiveness of \model~using POI prediction and friend suggestion tasks and show that it outperforms previous state-of-the-art POI recommendation methods. Our investigation shows that reason for the improvement stems from the process of (i) heterogeneous graph construction, and (ii) negative sampling. We show that the processes used in previous methods are more suitable to OSN check-in data based on sparse voluntary reporting, than dense(r) check-in data based on location tracking. 

To summarize, our work makes the following contributions:
\begin{enumerate}
\item Presents the first educational ``check-in'' dataset and explores its unique mobility and social characteristics;
\item 
Identifies the challenges for time-aware POI prediction in educational check-in data based on increased density due to location-based tracking (compared to previous voluntary-report LSBN data);
\item Proposes a novel heterogeneous information network-based model to encode the relations between users, POIs, and time-slots, and evaluates its efficacy for POI and user recommendation tasks.
\end{enumerate}

\section{Data Characteristics}
In this section, we discuss the characteristics of the Purdue educational ``check-in'' dataset and showcase its unique aspects compared to previous check-in data.

\subsection{Data sample}
In this work, we use two sample datasets from the Purdue Office of Instituional Research: (i) WiFi log data and (ii) building location profiles\footnote{Collected and analyzed anonymously, with IRB approval.}. 
We consider a sample of the data restricted to freshmen students in the 2016-17 academic year.
The 376Gb WiFi log file contains over 1 billion entries, each of which records a data communication between a campus WiFi access point and a personal device in the time period 7/31/2016 to 6/30/2017. Each entry contains activity time (date, hour and minute), anonymized user id, MAC address, and building id. The building profile provides building information, including building id, name, category, and functionalities. Note that each building belongs to one category but might have multiple functionalities. 
We remove users with fewer than 100 check-ins. We also drop the check-in records generated by MAC addresses that only checkin at a single building, as these devices are likely to be stationary PCs in dorms or offices. Moreover, using class registration information, we attach a `in-class' label for the record if the WiFi access point is in the building associated with their course schedule at that time/day. While, we retain these in-class checkins for the analysis in this section, we remove them for the modeling in Section~\ref{sec:method}, to focus the prediction task on less predictable user movements.  The final processed sample has 540 million logs in total. More dataset statistics are shown in Table~\ref{tab:data}.

\begin{table}
\begin{tabular}{|c|p{0.15\columnwidth}|p{0.45\columnwidth}|}
\hline
 {\em Item} &  {\em Number} & {\em Description} \\ \hline
 Users &  6250 & Freshmen \\ \hline
 POIs &  221 & On-campus buildings \\ \hline
 POI category  &  4  & Academic, Residential, Administration, Auxiliary\\ \hline
 POI functionality &7& Residence, Recreation, Dining, Exercise, Library/Lab, Classrooms, Others \\ \hline
 Time span & 1 sem. & Fall: 08/22/16 to 12/17/16 \\
\hline

\end{tabular}
\caption {Dataset description}
\label{tab:data}
\vspace{-6.mm}
\end{table}

\subsection{Temporal dynamics of user preferences}

\begin{figure*}[htp]
 \hspace*{-.2cm} 
\subfloat[Weekday]{
    \includegraphics[height = 3.5cm,trim=0 0 60 0, clip]{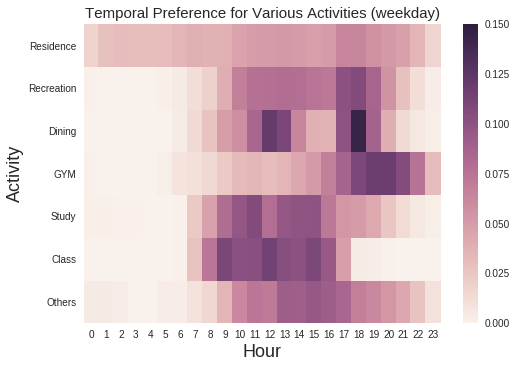}
    \label{fig:subweekday}
}
\subfloat[Weekend]{
    \includegraphics[height = 3.5cm,trim=20 0 60 0, clip]{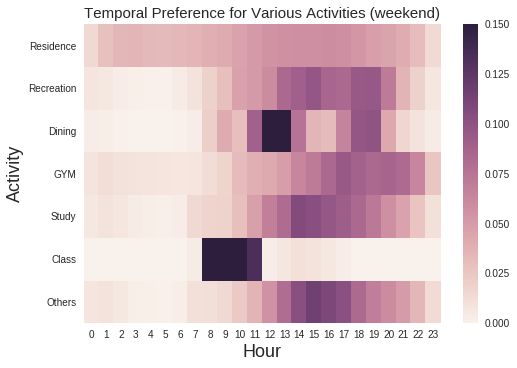}
    \label{fig:subweekend}
}
\subfloat[CS]{
    \includegraphics[height = 3.5cm,trim=20 0 60 0, clip]{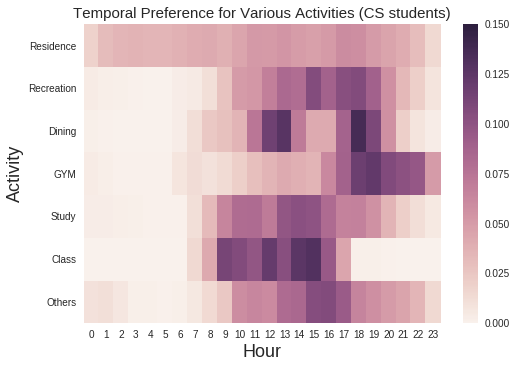}
    \label{fig:subcs}
}
\subfloat[Pharmacy]{
    \includegraphics[height = 3.5cm,trim=20 0 0 0, clip]{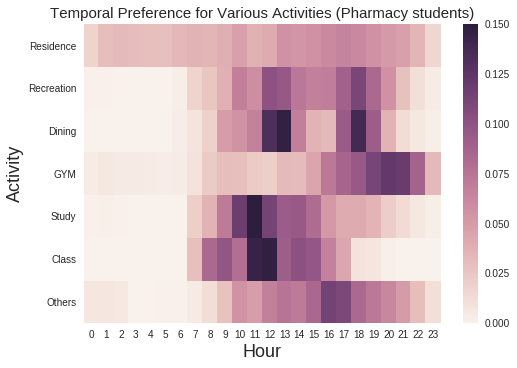}
    \label{fig:subphar}
}
\vspace{-3.mm}
\caption[Optional caption for list of figures]{Hourly activity preference for \subref{fig:subweekday} weekday, \subref{fig:subweekend} weekend, \subref{fig:subcs} computer science students, and \subref{fig:subphar} pharmacy students }
\label{fig:activity}
\end{figure*}

Figures \ref{fig:subweekday}-\ref{fig:subphar} show the students aggregated temporal preference for each type of activity in terms of the conditional probability $Pr(time = \tau|activity = a)$ for a given time slot $\tau$ and activity $a$ (e.g., Dining). We can see that different activities show unique temporal patterns. For example, on weekdays (Fig.~\ref{fig:subweekday}) students usually visit the dining halls (i.e., dining activity) around 12pm and 6pm, and go to the gym around 8pm. Check-ins at the residence halls are visible throughout the day, reflecting the variability in students daily routines and the dorms' versatility. 

Figures \ref{fig:subweekday} and \ref{fig:subweekend} show the differences in time preferences for weekday and weekend, respectively.  Students only have classes on Saturday morning, and they are more likely to start studying (including staying in the lab or library) at  later times on weekends. For non-academic activities on weekends, visiting hours to the gym are more distributed owing to their more flexible schedule, and more students choose to have lunch rather than dinner on campus. 

We also investigate if students' temporal preferences vary by major. Figures \ref{fig:subcs} and \ref{fig:subphar}, show the preferences for 302 computer science students and 267 Pharmacy students, respectively. We can see that the overall preferences are similar for $Dining$, while there are some differences in taking classes (shown in activity $Class$), staying in research labs/library (shown in activity $Study$), and exercise (shown in activity $Gym$). Specifically, Pharmacy students attend class more often from 11am to 12pm, while CS students attend class from morning to afternoon. CS students spend more time in academic buildings (from 10am to 7pm) than Pharmacy students who prefer to study in the morning and around noon. For non-academic activities, students from both majors show similar temporal preference, while pharmacy students tend to go to the gym at later times. 

\subsection{Co-visitation behavior}\label{sec:covisit}
While individual visitation histories can indicate temporal and spatial preference, in isolation they do not indicate relationships among the students. However, {\em co-visitation} events (i.e., when two students are in the same place at the same time), may be a noisy indicator of relations among students. Any one co-visitation event may be due to random chance, but a larger set of events, particularly when `in-class' events are dropped, is likely to indicate student friendship. To the best of our knowledge, a study of user pairwise co-visitation events hasn't been investigated in other spatio-temporal analyses. 

Since our dataset contains discrete WiFi login records, we merge each user's consecutive logins in the same building, 
and assume that the user stays in the building throughout this period. For example, if a user checked in at the library every four to eight minutes from 5pm to 6pm with no checkins at other buildings in between, we will merge these check-ins and record that the user stayed in library from 5pm to 6pm. In this way, we augment the visit history with duration time for each user, and use that to compute the pairwise co-visitation count matrix. As each co-visitation is per minute, the pairwise co-visitation count is the total time (in minutes) two users spend together in the same building. 
For example, the pair of users with the largest co-visitation count spent 38,129 minutes together, which is roughly 27 days, more than 25\% of the semester. Note that the in-class check-ins are removed for computing co-visitation. In this way, we only consider the activity outside of class for analyzing co-visitation, which we believe is more informative for determining friend realtions.
We will use the pairwise co-visitation count to examine the performance of our embeddings on a friend suggestion task in Section~\ref{sec:friend}.

\begin{figure}[th!]
\centering
     \includegraphics[width=6.5cm]{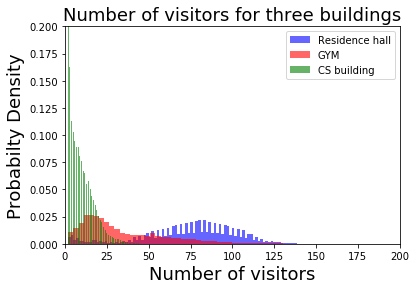}
\vspace{-3.mm}
\caption{Histogram of co-visitation size for an academic building (CS), the Gym, and a residence hall, over all times.}
\label{fig:visitor}
\end{figure}

Once the visit history is augmented with duration time for each user, we compute the number of users visiting  the same building at the same time. For each building, we calculate the number of unique visitors for each minute over the semester and filter out the moments when there are fewer than two visitors. We show the normalized histogram in Figure \ref{fig:visitor}. It indicates that the number of users appearing at the same time in the building may reflect building categories, as co-visitation happens more frequently in dorms and the gym compared to academic building (e.g., CS).

\subsection{Exploration behavior}

\begin{figure*}[htp]
\subfloat[Student ``check-in'' dataset]{
    \includegraphics[width = 5.5cm]{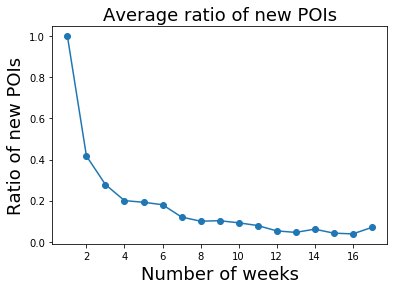}
    \label{fig:explore ours}
}
\subfloat[Student ``check-in'' dataset]{
    \includegraphics[width = 5.5cm]{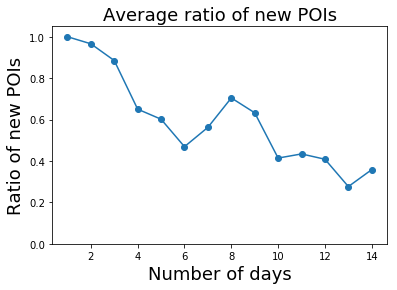}
    \label{fig:explore ours_1}
}
\subfloat[Foursquare and Gowalla]{
    \includegraphics[width = 6.30cm,trim=10 0 0 0, clip]{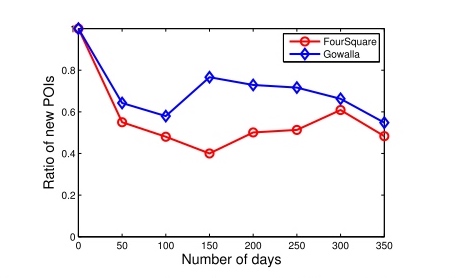}
    \label{fig:explore others}
}
\vspace{-2mm}
\caption{Average ratio of new POIs: (a-b) Purdue data (weeks/days), and (c) foursquare and Gowalla dataset (from~\cite{feng2015personalized}).}
\label{fig:explore}
\end{figure*}

We compare the exploration behaviors in our educational ``check-in'' dataset to traditional POI recommendation datasets like Foursquare and Gowalla. Figure~\ref{fig:explore} shows the average ratio of new POIs over all users for every new week. For example, the ratio at week two is the proportion of POIs visited during the second week that have not been visited in previously.

Compared with Foursquare users (Figure~\ref{fig:explore others}) who keep exploring new POIs all year round, freshmen (Figure~\ref{fig:explore ours}) appear to explore the campus very quickly (within 2-3 weeks), and then stick to a fixed range of buildings over the remainder of the semester. But when we zoom into the first two weeks (Figure \ref{fig:explore ours_1}), new students show similar exploration behaviors as in the Foursquare data, with 40 to 60 percent new POIs every day. This provides us with a unique opportunity to model two types of behaviors with different slices of the data: (1) the first few weeks of freshmen semester---exploring new places, and (2) the latter half of the semester---routinely visiting familiar places in a relatively limited activity range. 
\section{Proposed \model~Method} \label{sec:method}
In this section, we outline our proposed heterogeneous graph embedding method for POI prediction. Specifically, we consider a time-aware location prediction problem. Given a user and time slot (e.g., Monday 8 am), the model should predict a place that is most likely to be visited. 

We refer to our method as {\em Embedding for Dense Heterogeneous Graphs} (\model). It is designed specifically to reflect the characteristics of our educational check-in data, which is more dense than traditional LBSN check-in data. 
To better leverage contextual information, we propose a joint embedding model, which maps user, location, time and activity category into a common latent space. 

In this section, we introduce \model~step by step. We first construct a heterogeneous graph using the check-in records, then we learn continuous feature representations for vertices by capturing features of connectivity and structural similarity for pairs of nodes. In Sections~\ref{sec:poi-predict}-\ref{sec:friend-predict}, we discuss how to use the learned representations for POI prediction and friend suggestions, respectively. 


\subsection{Heterogeneous Graph Construction} \label{sec:graph-construct}

\noindent \textbf{Time indexing scheme.}
According to our data exploration results, the temporal characteristics of students behavior contain two aspects: (1) periodicity, and (2) preference variance. For example, students' check-ins have clear weekly cyclic patterns. Moreover, students usually visit academic buildings more on weekdays and stay at resident halls more on weekends. 

In order to capture these temporal cyclic patterns, we designed a time indexing scheme to encode a standard time stamp to a particular time id. We consider the preference variance in two scales: hours of a day and different days of a week. 
%
First, a time stamp is divided into two slices in terms of weekday and hour slot. Next, we split a week into 7 days and a day into the following four sessions: 
\begin{enumerate}
\item Morning -- hours between 6 am and 11:59 am
\item Afternoon -- hours between 12 pm and 4:59 pm
\item Evening -- hours between 5 pm and 11:59 pm
\item Night -- hours between 12 am and 5:59 am
\end{enumerate}
This totals 28 distinct time slot ids, which can represent both weekly and daily preference variance. 

\begin{figure}[th!]
\centering
    \includegraphics[width=6.5cm]{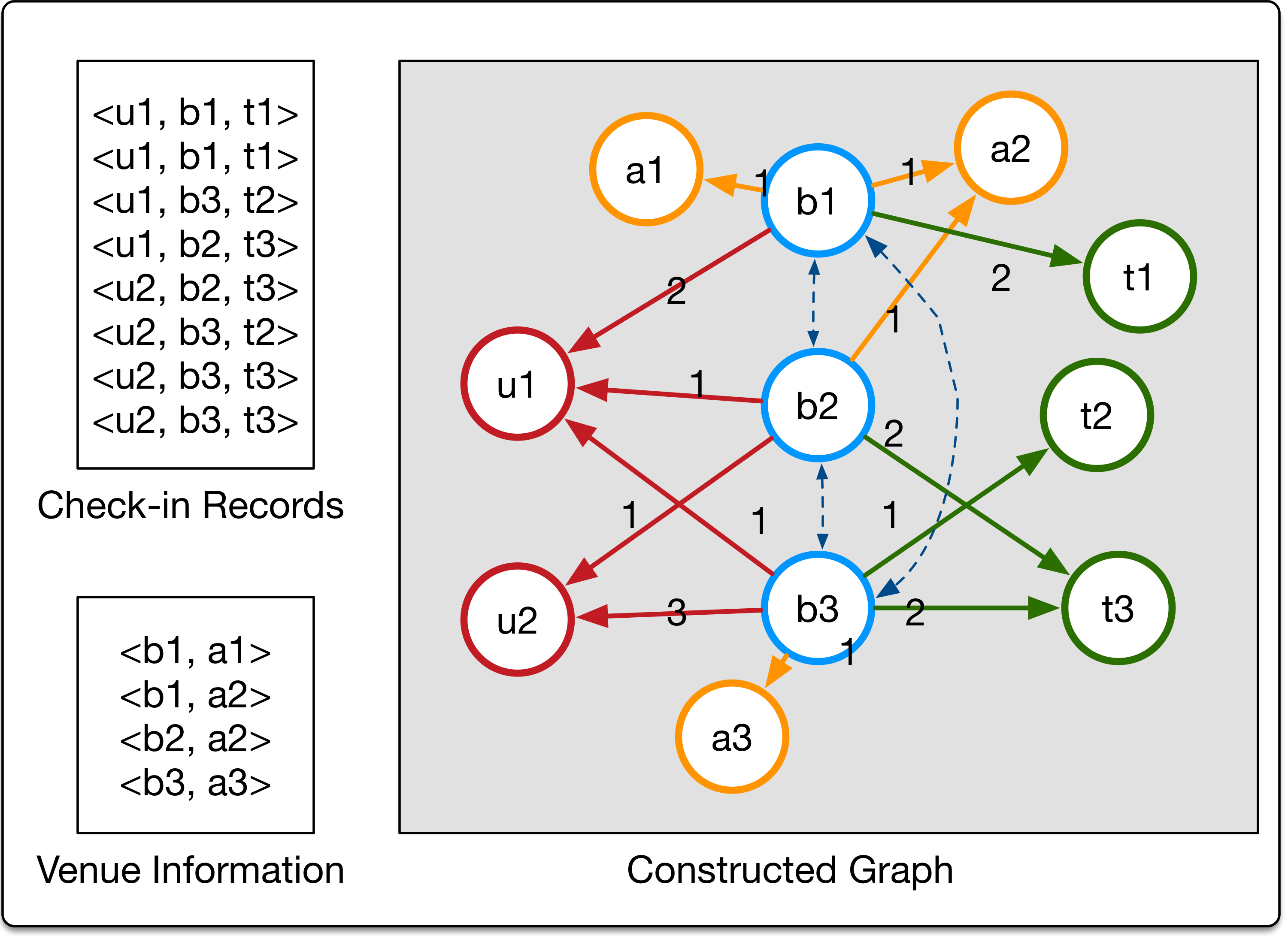}
\vspace{-1mm}
\caption{Heterogeneous graph constructed using eight example check-in records and venue information}
\label{fig:graph-sample}
\vspace{-2.mm}
\end{figure}

\vspace{2.mm}
\noindent \textbf{Weighted graph construction.}
We construct a weighted heterogeneous information network 
by aggregating the check-in records and venue information. An example is shown in Figure \ref{fig:graph-sample} with eight check-in records. In this example, $u_1, u_2$ denote two users, $b_1, b_2, b_3$ denote three buildings/POIs, $t_1, t_2, t_3$ denote three time slots, and $a_1, a_2, a_3$ are three types of activities corresponding to POI functionalities, which we obtained from the venue information. Our model considers three types of edges, i.e., POI-user, POI-time and POI-activity. For POI-time and POI-user edges, edge weights are co-occurrence counts for pair of nodes in the check-in records. For POI-activity edges, edge weights are set to 1. Our constructed graph contains 6250 user nodes, 221 POI nodes and 7 activities nodes. The POI-user graph density is 22.45\%, the POI-time graph density is 82.35\%, and the POI-activity graph density is 17.19\%. 

Note that past work \cite{xie2016learning} has included POI-POI edges in the graph by considering user transitions from one POI to another. However, these edges increase the average density of our graph substantially (POI-POI density: 60.13\%). As we will show in the experiments, these edges degrade the performance of the model, particularly on unvisited nodes, so we do not include them in the graph.

\vspace{-1mm}
\subsection{Graph Embedding}
We adapt the graph embedding approach from ~\citet{xie2016learning}, which is an extension of \cite{tang2015pte} geared for POI recommendation. These approaches are all based on the skip-gram model \cite{mikolov2013distributed} applied to graphs. Given an instance (word/node) and its context (neighbors), the objective of skip-gram is to minimize the log loss of predicting the context using the instance embedding as input features.
We employ a similar objective (as described below), but adjust the negative sampling approach to better fit the characteristics of the heterogeneous graph in our setting.


Specifically, we partition our heterogeneous graph into three bipartite graphs (POI-user graph $G_{bu}$, POI-time graph $G_{bt}$ and POI-activity graph $G_{ba}$). Below, we first introduce the graph embedding method for each bipartite graph, then we present our approach for negative sampling, and finally we show how to jointly learn the embeddings over the whole graph. 

\vspace{2.mm}
\noindent \textbf{Bipartite graph embedding.}
%
Given a bipartite graph $G_{AB} = (V_A\cup V_B, E)$ where $V_A$ and $V_B$ are two disjoint sets of vertices of different types, and $E$ is the set of edges between them, our task is to find the parameters $\theta$ of a model $p_\theta(v_i|v_j)$ ($v_i \in V_A$: context vertex; $v_j \in V_B$: target vertex) that closely approximates the empirical distribution $\tilde{p}(v_i|v_j)$ in terms of minimizing cross-entropy. Here the empirical distribution is given by the graph, i.e., 

\vspace{-2mm}
\begin{equation*}
\tilde{p}(v_i|v_j) = \frac{w_{ij}}{deg(j)}
\end{equation*}

\vspace{-0mm}
\noindent 
where $w_{ij}$ is the edge weight between $v_i$ and $v_j$, or zero if $v_i$ and $v_j$ are not connected. 

We define the conditional probability of vertex $v_i$ generated by vertex $v_j$ as the outcome of a softmax function:

\vspace{-2mm}
\begin{equation}
p_\theta(v_i|v_j) = \frac{e^{\vec{z_i}^T\vec{z_j}}}{\Sigma_{i'\in V_A}{e^{\vec{z_{i'}}^T\vec{z_j}}}}
\label{eq: distribution}
\end{equation}

\vspace{-0mm}
\noindent where $\vec{z_v}$ denotes the embedding for a vertex $v_v$.
For each vertex $v_j$ in $V_B$, Eq.\ref{eq: distribution} defines a conditional distribution $p(\cdot|v_j )$ over all the vertices in the set $V_A$. 
For each pair of vertices $v_j$, $v_{j'}$, their second-order proximity can actually be determined by their conditional distributions $p_\theta(\cdot|v_j )$, $p_\theta(\cdot|v_{j'})$. 

To learn embeddings that ensure the conditional distribution $p_\theta(\cdot|v_j)$ closely approximates the empirical distribution $\tilde{p}(\cdot|v_j)$, we minimize the following objective function over the graph $G_{AB}$:

\vspace{-2mm}
\begin{equation}
O_{AB}  = \sum_{j\in V_B}{\lambda_j d\!\left(\tilde{p}(\cdot|v_j), p_\theta(\cdot|v_j )\right)}
\end{equation}

\noindent where $d(\cdot, \cdot)$ is the KL-divergence between two distributions,  and $\lambda_j$ is the importance of vertex $v_j$ in the graph. Replacing $d(\cdot, \cdot)$ with KL-divergence, setting $\lambda_j$ = $deg(j)$ = $\sum_{i\in V_A}{w_{ij}}$ and omitting some constants, the objective function can be written as:

\vspace{-2mm}
\begin{equation}
O_{AB}  = -\sum_{(i, j)\in E}{w_{ij}\log{p_{\theta}(v_i|v_j)}}
\label{eq:obj}
\end{equation}

\noindent \textbf{Negative sampling.}
Optimizing the objective in Eq.~\ref{eq:obj} is computationally expensive,
as it requires the summation over the entire set of vertices when calculating the conditional probability $p_{\theta}(\cdot|v_j)$. 
To address this problem, we adopt the approach of negative sampling proposed in word2vec~\cite{mikolov2013distributed}, which instead of considering all pairs of nodes, samples a smaller set of observed edges, and then samples multiple ``negative'' edges for each observed edge. Specifically, in each step, a binary edge $e = (i, j)$ is sampled with the probability proportional to its weight $w_{ij}$, and then multiple negative edges $(i', j)$ are sampled from a specified noise distribution $q(i')$. 

The default noise distribution used in word2vec (and subsequently used by most, if not all, skip-gram based graph embedding models) is defined as a unigram distribution: $q(i) \propto deg(i)^{3/4}$, where $deg(i)$ denotes the degree of vertex $v_i$. This means that more ``popular'' vertices are more likely to be selected as negative samples. This makes sense in most NLP and graph embedding problems, where the word co-occurrence matrix or graph adjacency matrix is very sparse. The intuition behind this form of negative sampling is to distinguish between the true context word/vertex and another popular word/vertex which is unlikely to be a context. 

However, the graph adjacency matrix is relatively dense in our WiFI check-in data, due to longer user trajectories (i.e., more frequent check-ins). For example, our POI-POI graph adjacency matrix density is 60.13\%, whereas in the foursquare dataset the POI-POI graph is extremely sparse with 0.03\%  density. If we use the above popularity-based negative sampling method for our data, we find that 96\% of POI vertices sampled as ``negatives'' are actually connected to the target vertices---which obviously hinders estimation. 

To address this issue, we define a new process for efficient negative sampling utilizing the global statistics, i.e., the graph adjacency matrix. Moreover, we integrate the POI categorical information into the noise distribution. When a POI is from a popular category, it's less likely to be a {\em true} negative sample, i.e., it's more likely to be connected to the target vertex. By incorporating the global statistics and POI categorical information into the negative sampling procedure, our \model~model incorporates global features into the local predictive method. In practice, we replace the default noise distribution $q(i)$ with alternative $q(i|j)$. Here $v_j$ is the  given target vertex, and $v_i$ is the  generated negative sample vertex: 

\vspace{-2mm}
\begin{equation}
q(i|j) \propto 1 - \frac{w_{ij}}{deg(i)}\times Pr(cat(i))
\end{equation}

\noindent where $w_{ij}$ denotes the weight of edge $e_{ij}$, or equals zero if there is no edge between vertex $v_i$ and $v_j$, and $deg(i)$ is $v_{i}$'s degree. $Pr(cat(i))$ is the ratio of checkins in POIs with same category as POI $i$, or equals 1 when vertex $i$ is not a POI node. Note that $cat(i)$ corresponds one of the four POI categories in Table~\ref{tab:data}. 


Using edge sampling as in~\cite{tang2015line} and negative sampling as described above, our final objective function for the bipartite graph $G_{AB}$ is: 
\begin{equation}
O_{AB} = -\Sigma_{(i, j)\in E}
\left[\log{\sigma(\vec{z_i}^T\vec{z_j})} + \Sigma_{n=1}^{m}{E_{v_{i'} \sim q(\cdot|j)} \left(\log{\sigma(-\vec{z_{i'}}^T\vec{z_j})}\right)}\right]
\label{eq:final obj}
\end{equation}

Here $\sigma$ refers to the sigmoid function and we sample $m$ negative examples for each positive example.
In our implementation, we use the alias table method from \citet{li2014reducing} to draw a negative sample with a pre-computed alias table based on the noise distribution $q(\cdot|j)$. This ensures that it takes O(1) time to repeatedly draw samples from the same distribution. In this way we can achieve the same time complexity as the original LINE model, which is demonstrated to be scalable. 
Then we adopt the asynchronous stochastic gradient algorithm (ASGD) \cite{recht2011hogwild} to optimize Eq. \ref{eq:final obj}. In each iteration, if the edge $e_{ij}$ is sampled, the gradient w.r.t. the embedding vector $\vec{z_i}$ of vertex $v_i$ will be calculated as $\frac{\partial O_{AB}}{\partial \vec{z_i}}$.

\vspace{2.mm}
\noindent \textbf{Joint training.}
The overall objective is the sum of the objectives for three bipartite graphs $G_{bu}$, $G_{bt}$ and $G_{ab}$:

\vspace{-2mm}
\begin{equation}
O = O_{bu} + O_{bt} + O_{ab}
\end{equation}

\noindent where each component objective $O_{bu}$, $O_{bt}$ and $O_{ab}$ is specified by Eq. \ref{eq:final obj}. 
We learn a joint node embedding by iterating through the three component bipartite graphs in a round-robin fashion and updating the vector representations in each bipartite graph embedding procedure. See Algorithm~\ref{alg:train} for more details.

\begin{algorithm}
\caption{\model~training algorithm}\label{alg:train}
\textbf{Input:} Bipartite graphs (POI-user graph $G_{bu}$, POI-time graph $G_{bt}$, POI-activity graph $G_{ab}$), number of iterations $N$, negative sample size $m$, vector dimension $d$.\\

\textbf{Output:} latent node embeddings for---users: $Z_u \in R^{|U|\times d}$, POIs: $Z_b \in R^{|B|\times d}$, time slots:  $Z_t \in R^{|T|\times d}$, and activities: $Z_a \in R^{|A|\times d}$.
\begin{algorithmic}[1]
\Procedure{Joint train}{$N, m, d, G_{bu}, G_{bt}, G_{ab}$}
\State Initialize $Z_u$, $Z_b$, $Z_t$ and $Z_a$
\While{$iter\leq N$}
\State \textbf{Bipartite graph embedding}($G_{bu}, m$) \\ \Comment{update $Z_b$, $Z_u$}
\State \textbf{Bipartite graph embedding}($G_{bt}$, m) \\ \Comment{update $Z_b$, $Z_t$}
\State \textbf{Bipartite graph embedding}($G_{ab}$, m) \\ \Comment{update $Z_a$, $Z_b$}
\EndWhile
\State \textbf{return} $Z_u$, $Z_b$, $Z_t$ and $Z_a$
\EndProcedure
\end{algorithmic}
\begin{algorithmic}[1]
\Procedure{Bipartite graph embedding}{$G_{AB}, m$}
\State sample an edge $e_{ij}$ ($v_i \in V_A, v_j \in V_B$) 
\State sample $m$ negative nodes from $q(\cdot|j)$ (denote as $v_{i'}$)
\State update $z_i$, $z_j$, and $z_{i'}$ to minimize Eq. \ref{eq:final obj}.
\EndProcedure
\end{algorithmic}
\end{algorithm}

\vspace{2mm}
\subsection{Predicting POIs using Embeddings}\label{sec:poi-predict}
Once we have trained our model and learned representations for users, time slots, and locations, we can perform location prediction on new check-in data using simple operations on vectors. Given a query $(user, time)$ i.e., $q = (u, \tau)$, we first project the timestamp $\tau$ into time slot $t$ using the time indexing scheme described in Section~\ref{sec:graph-construct}, and then rank the POIs based on their location in the embedding. More precisely, given a query $q = (u, \tau)$, for each POI $b$, we compute its ranking score as: 

\vspace{-1mm}
\begin{equation}
S(b \: | \: u, \tau=t) = \vec{z_b}^T\vec{z_u} + \vec{z_b}^T\vec{z_t} 
\end{equation}

\noindent where $\vec{z_b}, \vec{z_b}, \vec{z_t}$ are embeddings for user $u$, POI $b$, time slot $t$ respectively. Then we select the $k$ POIs with the highest ranking scores as predictions. 
Note the POI embedding $\vec{b}$ reflects activity information via the POI-activity graph, since our model jointly learns the embedding of multiple relational networks in the same latent space. Therefore, for both visited POIs and unvisited POIs (also called cold-start POIs), we can perform user recommendations using the same scoring function. 

\subsection{Suggesting Friends using Embeddings}\label{sec:friend-predict}
As the embeddings learned from the model fuse the interactions between user-POI, POI-time and POI-activity, we can make use of the embeddings to suggest potential friends for a given user based on their pairwise similarity. Specifically, for a query user $u$, $\forall v\in U\setminus u$, we compute ${z_u}^T z_v$ and rank the results over $U$, the set of users. From this, we return the top ranked users as people that are more likely to be friends of $u$. 



\vspace{2mm}
\section{Experimental Evaluation}
\subsection{Methodology}
In the experiments, we concatenate each student's first 80\% check-in records in chronological order to create the training set examples and then use the remaining 20\% as the test set. 
We set the number of iterations ($N$) to 100M with a batch size of 1, the dimension of the embedding vector ($d$) is set to 100, and we sample 10 negative samples ($m$) for each vertex pair.

We use accuracy@k as the measure of prediction effectiveness, which is a commonly used metric for this task (see e.g., \cite{feng2015personalized}, \cite{xie2016learning}). However, in contrast with previous work, which only compare the score of the true POI to the score of {\em unvisited} POIs during evaluation, we evaluate by comparing the true POI's score to the score of all other POIs (both visited and unvisited). 
Specifically, for each check-in record (user, time, POI) in the testset, we recommend the top $k$ POIs for the query (user, time) as described in \ref{sec:poi-predict}, and determine if the true POI appears in the top-k list (which is defined as a 'hit'). The accuracy@k is defined as the ratio of hits to the testset size.

\subsection{Comparison Models}
We compare our proposed model \model~ to baselines, state-of-the-art alternative methods and \model~ variants.

\vspace{2mm}
\noindent \textbf{NBC}: Naive Bayes classifier using (user, time-slot) as joint features. For each query $(u, t)$, the probability of predicting POI $b$ is given by $p(b|u, t) \propto p(u,t|b)\cdot p(b)$ where $b$ denotes a candidate POI, and $(u, t)$ denote a (user, time-slot) pair. This is a strong baseline which takes into account POI popularity and a combination of personal and temporal preference based on counting.

\vspace{2mm}
\noindent \textbf{GE} \cite{xie2016learning}: The state-of-art graph embedding method for time-aware POI recommendation (developed using Foursquare and Gowalla data). GE uses POI-POI edges, POI-time edges, POI-region edges, and POI-activity edges, and jointly embeds POIs, times, regions and activities into a latent space. User embeddings are computed as sum of recent visited POIs' embeddings. See Section~\ref{sec:related} for details.  

\vspace{2mm}
\noindent \textbf{GE++}: An augmented version of GE that we create to assess the effect of learning user embeddings directly during joint training. This version of GE incorporates POI-user edges in the graph, in addition to its heterogeneous graph embedding. 

\vspace{2mm}
\noindent \textbf{\model}: Our proposed model, where we include the POI-user graph, the POI-time graph, and the POI-activity graph with our improved negative sampling method.

\vspace{2mm}
\noindent \textbf{\model-NS}: A simplified version of \model~, in which we use the traditional method for generating negative samples based on vertex degree. 

\vspace{2mm}
\noindent \textbf{\model-POI}: An augmented version of \model, where we also include the POI-POI bipartite graph in the heterogeneous graph for learning the embeddings. Note that we only record a POI-POI edge is there is a transition between the two POIs within a four hour time window.

\vspace{2mm}
All the models are run on a single machine with 8G memory using 20 threads. Both \model~ and its variants are very efficient---it takes about 18 minutes (excluding pre-computation of negative sampling alias table) to process a network with 6486 nodes and 315,407 edges.

\subsection{Predictive Effectiveness}
\begin{figure}[th!]
\centering
    \includegraphics[width = 7cm]{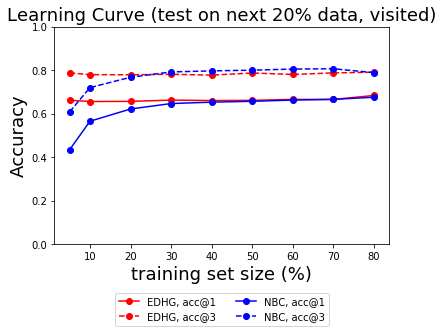}
\vspace{-2.mm}
\caption{Learning curve for visited POIs}
\label{fig:visited}
\vspace{-4.mm}
\end{figure}

\begin{figure}[th!]
\centering
    \includegraphics[width = 7cm]{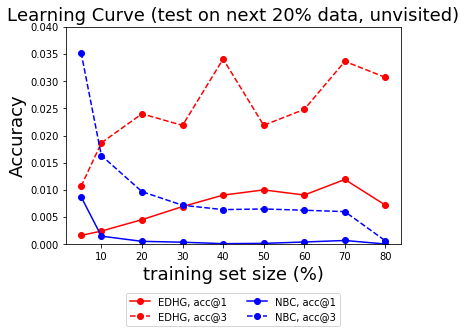}
\vspace{-2.mm}
\caption{Learning curve for unvisited POIs}
\label{fig:unvisited}
\vspace{-2.mm}
\end{figure}

Here we present the experimental results for all prediction methods using well-tuned parameters. 
Prediction effectiveness in terms of accuracy@k is shown in Table~\ref{tab:accuracy}. We report results for visited and unvisited POIs to highlight the difference between in-sample and out-of-sample performance. We also use the 20\% test data to show learning curves for accuracy@1 and @3 in Figures~\ref{fig:visited}, \ref{fig:unvisited} for visited and unvisited POIs respectively. 
From the results we can make the following observations:

\begin{table}
\hspace*{-0.2cm} 
\begin{tabular}{|c|l|cccc|}
\hline
 Type & \diagbox{Model}{Acc@k} & k = 1 & k = 3 & k = 5 & k = 10 \\
\hline
 & GE & 0.1079 & 0.3781 & 0.5104 & 0.6543\\
&GE++ & 0.3019 & 0.5190 & 0.6063 & 0.6909\\
visited &\model-NS & 0.3321 & 0.5846 & 0.7024 & 0.8137\\
&\model-POI & 0.6832 &  0.7912 & 0.8368 & 0.8954\\
&\model & \textbf{0.6846} & \textbf{0.7915} & 0.8367 & 0.8961\\
& NBC & 0.6765 & 0.7895 & \textbf{0.8495} & \textbf{0.9016} \\
\hline
 & GE & 0.0027 & 0.0073 & 0.0241 & 0.0641 \\
& GE++ & \textbf{0.0084} & 0.0227 & 0.0332 & 0.0671 \\
un- &\model-NS & 0.0057 & 0.0128 & 0.0301 & 0.0598 \\
visited  &\model-POI & 0.0034 & 0.0145 & 0.0195 & 0.0334 \\
 &\model & 0.0072 & \textbf{0.0307} & \textbf{0.0360} & \textbf{0.0710} \\
 & NBC & 5.4e-05 & 6.3e-04&0.0025 & 0.0133 \\
 \hline
  & GE & 0.1084 & 0.3720 & 0.5026 & 0.6443\\
 & GE++ & 0.2981 & 0.5125 & 0.5988 & 0.6828\\
total &\model-NS & 0.3270 &0.5772 & 0.6937 & 0.7996 \\
  &\model-POI &0.6744 & 0.7811 & 0.8261 & 0.8842 \\
 &\model& \textbf{0.6760} & \textbf{0.7816} & 0.8263 & 0.8854 \\
 & NBC & 0.6677 & 0.7793 & \textbf{0.8385} & \textbf{0.8901} \\
 \hline
\end{tabular}
\caption {Prediction accuracy}
\label{tab:accuracy}
 \vspace{-3.mm}
\end{table}

\vspace{2mm}
\noindent \textbf{\model~v.s. \model-NS}: the full \model~consistently outperforms \model-NS for both visited and unvisited POIs, with a 100\% performance gain in terms of accuracy@1, and 35.4\% in terms of accuracy@3. The significant performance gain is due to the improved negative sampling procedures, which selects more informative negative samples for SGD updates. This indicates that it is promising to customize the empirical noise distribution used in negative sampling for various tasks or datasets. 

\vspace{2mm}
\noindent  \textbf{\model~v.s. \model-POI}: The \model-POI variant includes the POI-POI transition graph in the original graph which prior work on GE claimed as an important component, but it doesn't  improve performance on our recommendation task, and it even downgrades performance for unvisited POIs. This indicates that transition behavior is not informative in our data, as there are too many transitions betweens buildings that cannot be explained by a single reason. 

\vspace{2mm}
\noindent  \textbf{\model~v.s. GE/GE++}: \model~significantly outperforms GE for both visited and unvisited POIs. The reasons might be due to (1) GE using the POI-POI graph to model the "locality" of individual check-ins for Foursquare data. However, as revealed by the comparison between  \model~and \model-POI, including the POI-POI graph doesn't help in our setting. Or (2) GE doesn't include users as entities in their graph representation, but computes the user embeddings based on recent visit histories. Due to the limited number of POIs in our data, computing the user embedding computed in this way may fail to capture personal preferences. Considering the performance of GE++, which we adapt to our data by adding the POI-user graph to the original GE model, modeling users in the graph helps improve its predictions, but the performance of GE++ is still inferior to that of \model.

\vspace{2mm}
\noindent \textbf{\model~v.s. NBC}: \model~achieves comparable performance for visited places and significantly outperforms NBC for unvisited places. In reality, 
when we look at the learning curves for prediction accuracy, Figure~\ref{fig:visited} shows that our model converges very fast while NBC needs more data to achieve a comparable result; and Figure~\ref{fig:unvisited} shows that our model actually ``learns'' how to recommend unvisited places with increasing accuracy, while NBC fails to deal with the cold-start recommendation problem, even when provided with a large amount of training data. 

\subsection{Parameter Sensitivity}
\begin{figure*}[htp]
\subfloat[Num. iterations v.s. Prediction accuracy]{
    \includegraphics[width = 5.5cm]{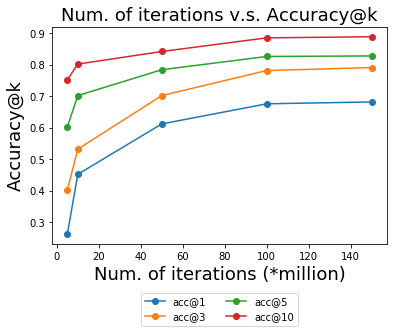}
    \label{fig:param: sample}
}
\subfloat[Vector dim. v.s. Prediction accuracy]{
    \includegraphics[width = 5.5cm]{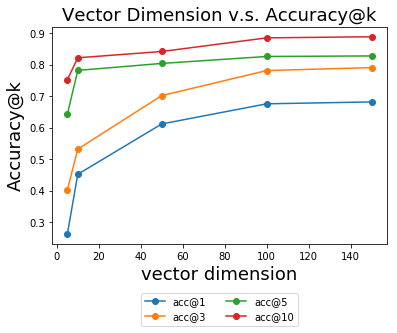}
    \label{fig:param: dim}
}
\subfloat[Negative sample size v.s. Prediction accuracy]{
    \includegraphics[width = 5.5cm]{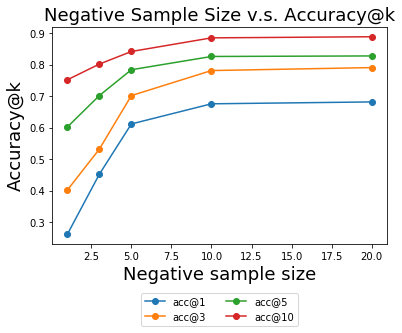}
    \label{fig:param: neg}
}
\vspace{-2.mm}
\caption[Optional caption for list of figures]{Impact of number of iterations, embedding vector dimension, and negative sample size on prediction accuracy@k.}
\label{fig:paramaters}
\end{figure*}

\noindent \textbf{Granularity of temporal pattern.}
In Table~\ref{tab:accuracy} we evaluated the predictive performance of our model with a combination of weekly of daily pattern using 28 time slots. Here, we design two additional variants to explore the effect of temporal patterns with different granularity. The \model-hour only considers time period of day (4 time slots) and \model-dow only considers day of week (7 time slots). The results are shown in Table \ref{tab:time}.

\begin{table}
\hspace*{-0.2cm} 
\begin{tabular}{|c|l|cccc|}
\hline
 Type & \diagbox{Model}{Acc@k} & k = 1 & k = 3 & k = 5 & k = 10 \\
\hline
 & \model-dow & 0.6025 & 0.7010 & 0.7544 & 0.8403\\
visited &\model-hour & 0.6727 & 0.7850 & 0.8284 & 0.8927\\
&\model & \textbf{0.6846} & \textbf{0.7915} & \textbf{0.8367} & \textbf{0.8961}\\
\hline
un- & \model-dow & 0.0041 & 0.0083 & 0.0141 & 0.0402 \\
visited &\model-hour & 0.0022 & 0.0080 & 0.0260 & 0.0490 \\
 &\model & \textbf{0.0072} & \textbf{0.0307} & \textbf{0.0360} & \textbf{0.0710} \\
 \hline
  & \model-dow & 0.5947 & 0.6920 & 0.7448 & 0.8299\\
total &\model-hour & 0.6639 &0.7727 & 0.8102 & 0.8821 \\
 &\model  & \textbf{0.6760} & \textbf{0.7816} & \textbf{0.8263} & \textbf{0.8854} \\
 \hline
\end{tabular}
\caption {Prediction accuracy v.s. temporal granularity.}
\label{tab:time}
\vspace{-3.mm}
\end{table}

From Table \ref{tab:time} we can see that both Day of Week and Hour of Day are important temporal factors. Specifically, when we only consider weekly patterns (day of week), prediction accuracy decreases by roughly 10\%; when we only consider daily patterns (hour of day), prediction accuracy slightly decreases by 1.5\%. This indicates that time-of-day effects are more significant than day-of-week effects in terms of POI prediction. It's likely that students, particularly freshman, have less flexibility in daily routines due to their course schedule, which makes time-of-day a more important factor for nearly all types of activities.

\vspace{2.mm}
\noindent \textbf{Number of iterations \& vector dimension.}
Figures~\ref{fig:param: sample} and \ref{fig:param: dim} show the performance of \model~with different number of iterations $N$ and embedding dimensions $d$. Note that, the units for $N$ is set to 1 million. We can see from Figure~\ref{fig:param: sample} that the accuracy increases and converges quickly when the number of iterations is larger than 50M. We used $N=100(M)$ to ensure convergence. For embedding dimension, we chose $d=100$ as the accuracy does not increase substantially after that point.

\vspace{2.mm}
\noindent \textbf{Negative sample size.}
Figure~\ref{fig:param: neg} presents the performance of \model~with different numbers of negative samples per example. With more negative samples the accuracy increases, and it plateaus when negative sample size is 10. Therefore, we chose $m=10$ negative samples for use during optimization.

\subsection{Friend Suggestion Effectiveness} \label{sec:friend}
To examine the efficacy of using \model's vector representation for suggesting friends, we first need to identify a proxy signal for evaluation (since we do not have ground truth information about friend relations among the students).
Specifically, we consider the following two ways data to determine ``true'' friends for evaluation:

\vspace{2.mm}
\noindent \textbf{Covisit} As in shown in Section~\ref{sec:covisit}, a co-visitation record is generated when two users check in at the same building at the same time (time unit: minute). In this approach, we identify ``friends'' of a query user as those with the largest co-visitation counts.

\vspace{2.mm}
\noindent \textbf{Location} Based on the user-building check-in count matrix, we create a ranking list of buildings for each user, with the most frequently visited building ranked highest. In this approach, we identify ``friends'' of a query user as those that have the smallest distance between the users' ranked list of buildings (using Kendall $\tau$ distance). 
We apply the friend suggestion to the most active users in our dataset, sorted by activity level. For each user, given a set of ``true'' friends from one of the baselines above, we evaluate the top 10 friend suggestions from \model~using Mean Reciprocal Rank (MRR). 
MRR is computed as: 

\vspace{-2mm}
\begin{equation*}
\frac{1}{|U|} \sum_{i = 1}^{|U|}{\sum_{j \in F_i}{\frac{1}{rank(j)}}}
\end{equation*}

\noindent where $U$ is the set of active users, $F_i=10$ is the set of ``true'' friends of user $i$ which are obtained from the data and $rank(j)$ is the rank of item $j$ in the ranking list. We compare the performance of \model~and GE in terms of MRR scores. 

Since the \textbf{Covisit} baseline encodes both temporal and geographical preference, and the \textbf{Location} baseline only takes into account geographical preference, \textbf{Covisit} is likely a better proxy for ``true'' friends, and thus the ideal search results should have a higher MRR score w.r.t. \textbf{Covisit}. 

\begin{figure}[th!]
\vspace{-2.mm}
\centering
	 \includegraphics[width = 7cm]{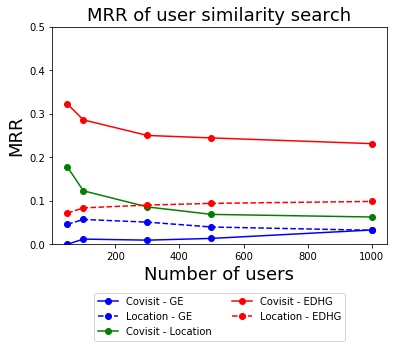}
\vspace{-2.mm}
\caption{Number of frequent users v.s. MRR scores}
\label{fig:mrr_new}
\vspace{-2.mm}
\end{figure}

The results are shown in Figure~\ref{fig:mrr_new}. We can make the following observations based on the results. 
Comparing \textbf{Covisit} and \textbf{Location}, \model~suggests friends with more relevance to co-visitation counts than merely geographical preference, while GE does the opposite. Since neither method uses user co-visitation data directly in its model, this implies that \model~captures social behaviors from the spatial-temporal data more accurately.
Comparing the general MRR scores of the two models, \model~suggests friends with higher accuracy in general. 
We also calculated the MMR between the \textbf{Covisit} and \textbf{Location} friend lists (the green line in the graph). The relatively low MRR score reveals that a large potion of co-visitation behavior cannot be explained by location preference. In reality, students with same major and same year usually stay in a same set of places, e.g. academic buildings and libraries, but their temporal preferences may vary significantly. 
The plot shows how performance changes as the increase of $|U|$. We can see that, on the co-visitation data, \model's MMR score decreases as less active users are included in $U$. This indicates that \model~discovers better suggestions for more active users, which suggest that with richer check-in information we can capture more precise social relationships.


\subsection{Visualization of Embeddings}

\begin{figure}[t!]
 \hspace*{-0.2 cm} 
\centering
    \includegraphics[width = 8.5cm]{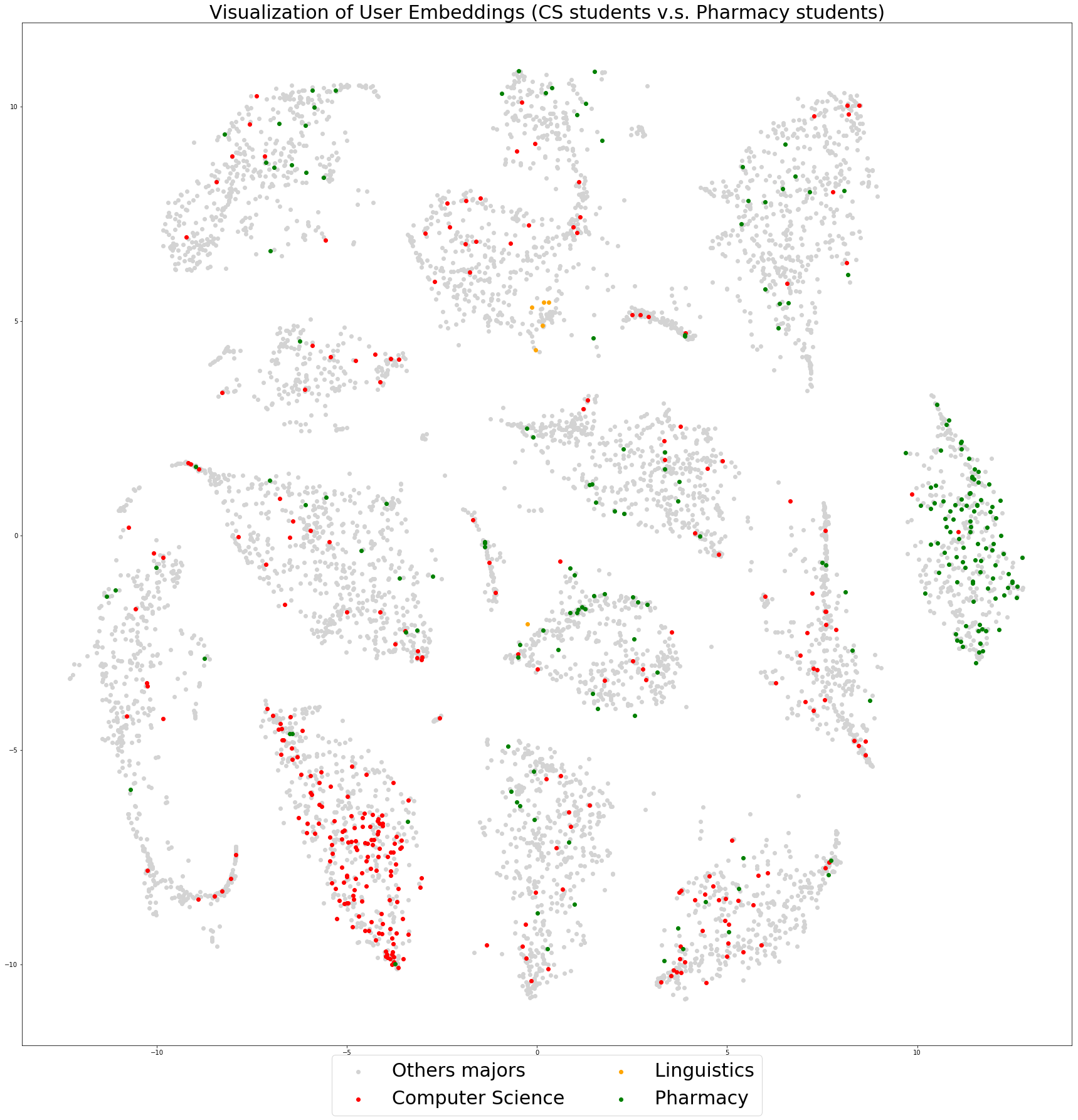}
\vspace{-3.mm}
\caption{User embeddings}
\label{fig:user_emb}
\vspace{-2.mm}
\end{figure}

Figure~\ref{fig:user_emb} shows a visualization of the learned user embeddings, where we project the $d=100$ dimensions into 2D using t-SNE \cite{maaten2008visualizing}. From the visualization we can clearly find two clusters of computer science students and pharmacy students (colored green and red respectively). This can be understood through their differences in temporal preference (as shown in Figures~\ref{fig:subcs}-\ref{fig:subphar}) and also their geographical preferences (i.e., these two majors share very few academic buildings). 

\section{Related Work}\label{sec:related}

POI recommendation methods have received extensive research attention in the last five years, and many approaches have been proposed. For example, Wang et al. (\cite{wang2015geo,wang2016spore}) applied sparse additive generative models to incorporate multiple factors for POI recommendation. \citet{yin2017spatial} proposed a Spatial-Aware Hierarchical Collaborative Deep Learning model (SH-CDL), which jointly performs deep representation learning for POIs and hierarchically additive representation learning for spatial-aware personal preferences. \citet{xie2016learning} proposed a graph embedding model GE for context-aware POI recommendation, which uses the POI-POI transition graph, POI-region graph, and POI-category graph, and jointly learn the representations for POI, region, and time with the same method as PTE \cite{tang2015pte}. User embeddings are then computed as weighted sum of recent POI embeddings, and with user, POI, time embeddings they can perform time-aware POI recommendation. GE achieved better performance than all previous work on this task, which is why we use it as a baseline for evaluation in this paper. 

In addition, our work is related to the extensive literature on network embedding, which has attracted a great deal of attention in recent years. Many of these recent methods are technically inspired by Skipgram \cite{mikolov2013distributed}. For example, Deepwalk \cite{perozzi2014deepwalk} uses the embedding of a node to predict the context in the graph, where the context is generated by a random walk. Metapath2vec \cite{dong2017metapath2vec} extends DeepWalk for heterogeneous graph embedding. LINE \cite{tang2015line} extends the skip-gram model to have multiple context spaces for modeling both first and second order proximity. PTE \cite{tang2015pte} adapts the LINE model \cite{tang2015line} for embedding bipartite networks. Note that PTE model is directly adopted by GE \cite{xie2016learning} for POI recommendation, and we further adjust PTE to our setting by improving the negative sampling method and targeting the graph construction process. 

Heterogeneous information network embedding has been broadly applied to multiple tasks. For example, \citet{tang2015pte} predicted text embeddings based on heterogeneous text networks which showed great potential in document classification. \citet{zhang2017react} proposed ReAct, a method that processes continuous geo-tagged social media (GTSM) streams into a heterogeneous graph and obtains recency-aware activity models on the fly, in order to reveal up-to-date spatiotemporal activities. \citet{chen2017task} proposed a task-guided and path-augmented heterogeneous network embedding for author identification task. 

\section{Conclusions}
This paper presents our analysis of the first educational ``check-in'' dataset and proposes \model, a heterogeneous graph embedding based method to model more dense spatio-temporal checkin activity. To account for the unique characteristics of the data, we improve the negative sampling method to incorporate global statistics of the graph/data into the noise distribution. We also show that it's better to drop the POI-POI transition edges when the check-in data is more dense. We evaluated \model~with two tasks: time-aware POI prediction and friend suggestion. On both tasks, our proposed model outperforms the previous state-of-art methods and baselines. These initial results indicate the promise of using student trajectory information for personalized recommendations in education apps, as well as in predictive models of student retention and satisfaction.   

Several interesting research problems remain for further exploration. For example, we did not make direct use of the co-visitation data in the model but rather withheld it for evaluation of the friend suggestions. We plan to incorporate it in the training process and see whether social interactions impact student checkin behavior. Also, inspired by~\citet{chen2018improving}, we may be able to further improve the negative sampling by dynamically selecting informative negative samples during each SGD update. 

\section*{Acknowledgements}
This research is supported by NSF under contract number(s) IIS-1546488 and IIS-1618690. The U.S. Government is authorized to reproduce and distribute reprints for governmental purposes notwithstanding any copyright notation hereon. The views and conclusions contained herein are those of the authors and should not be interpreted as necessarily representing the official policies or endorsements either expressed or implied, of NSF or the U.S. Government.

\bibliographystyle{ACM-Reference-Format}
\bibliography{7_bibliography}


\begin{thebibliography}{23}


\ifx \showCODEN    \undefined \def \showCODEN     #1{\unskip}     \fi
\ifx \showDOI      \undefined \def \showDOI       #1{#1}\fi
\ifx \showISBNx    \undefined \def \showISBNx     #1{\unskip}     \fi
\ifx \showISBNxiii \undefined \def \showISBNxiii  #1{\unskip}     \fi
\ifx \showISSN     \undefined \def \showISSN      #1{\unskip}     \fi
\ifx \showLCCN     \undefined \def \showLCCN      #1{\unskip}     \fi
\ifx \shownote     \undefined \def \shownote      #1{#1}          \fi
\ifx \showarticletitle \undefined \def \showarticletitle #1{#1}   \fi
\ifx \showURL      \undefined \def \showURL       {\relax}        \fi
\providecommand\bibfield[2]{#2}
\providecommand\bibinfo[2]{#2}
\providecommand\natexlab[1]{#1}
\providecommand\showeprint[2][]{arXiv:#2}

\bibitem[\protect\citeauthoryear{Chen, Yuan, Jose, and Zhang}{Chen
  et~al\mbox{.}}{2018}]%
        {chen2018improving}
\bibfield{author}{\bibinfo{person}{Long Chen}, \bibinfo{person}{Fajie Yuan},
  \bibinfo{person}{Joemon~M Jose}, {and} \bibinfo{person}{Weinan Zhang}.}
  \bibinfo{year}{2018}\natexlab{}.
\newblock \showarticletitle{Improving Negative Sampling for Word Representation
  using Self-embedded Features}. In \bibinfo{booktitle}{\emph{Proceedings of
  the Eleventh ACM International Conference on Web Search and Data Mining}}.
  ACM, \bibinfo{pages}{99--107}.
\newblock


\bibitem[\protect\citeauthoryear{Chen and Sun}{Chen and Sun}{2017}]%
        {chen2017task}
\bibfield{author}{\bibinfo{person}{Ting Chen} {and} \bibinfo{person}{Yizhou
  Sun}.} \bibinfo{year}{2017}\natexlab{}.
\newblock \showarticletitle{Task-guided and path-augmented heterogeneous
  network embedding for author identification}. In
  \bibinfo{booktitle}{\emph{Proceedings of the Tenth ACM International
  Conference on Web Search and Data Mining}}. ACM, \bibinfo{pages}{295--304}.
\newblock


\bibitem[\protect\citeauthoryear{Dong, Chawla, and Swami}{Dong
  et~al\mbox{.}}{2017}]%
        {dong2017metapath2vec}
\bibfield{author}{\bibinfo{person}{Yuxiao Dong}, \bibinfo{person}{Nitesh~V
  Chawla}, {and} \bibinfo{person}{Ananthram Swami}.}
  \bibinfo{year}{2017}\natexlab{}.
\newblock \showarticletitle{metapath2vec: Scalable representation learning for
  heterogeneous networks}. In \bibinfo{booktitle}{\emph{Proceedings of the 23rd
  ACM SIGKDD International Conference on Knowledge Discovery and Data Mining}}.
  ACM, \bibinfo{pages}{135--144}.
\newblock


\bibitem[\protect\citeauthoryear{Feng, Li, Zeng, Cong, Chee, and Yuan}{Feng
  et~al\mbox{.}}{2015}]%
        {feng2015personalized}
\bibfield{author}{\bibinfo{person}{Shanshan Feng}, \bibinfo{person}{Xutao Li},
  \bibinfo{person}{Yifeng Zeng}, \bibinfo{person}{Gao Cong},
  \bibinfo{person}{Yeow~Meng Chee}, {and} \bibinfo{person}{Quan Yuan}.}
  \bibinfo{year}{2015}\natexlab{}.
\newblock \showarticletitle{Personalized Ranking Metric Embedding for Next New
  POI Recommendation.}. In \bibinfo{booktitle}{\emph{IJCAI}}.
  \bibinfo{pages}{2069--2075}.
\newblock


\bibitem[\protect\citeauthoryear{He, Li, Liao, Song, and Cheung}{He
  et~al\mbox{.}}{2016}]%
        {he2016inferring}
\bibfield{author}{\bibinfo{person}{Jing He}, \bibinfo{person}{Xin Li},
  \bibinfo{person}{Lejian Liao}, \bibinfo{person}{Dandan Song}, {and}
  \bibinfo{person}{William~K Cheung}.} \bibinfo{year}{2016}\natexlab{}.
\newblock \showarticletitle{Inferring a Personalized Next Point-of-Interest
  Recommendation Model with Latent Behavior Patterns.}. In
  \bibinfo{booktitle}{\emph{AAAI}}. \bibinfo{pages}{137--143}.
\newblock


\bibitem[\protect\citeauthoryear{Kylasa, Kollias, and Grama}{Kylasa
  et~al\mbox{.}}{2016}]%
        {kylasa2016social}
\bibfield{author}{\bibinfo{person}{S Kylasa}, \bibinfo{person}{G Kollias},
  {and} \bibinfo{person}{A Grama}.} \bibinfo{year}{2016}\natexlab{}.
\newblock \showarticletitle{Social ties and checkin sites: connections and
  latent structures in location-based social networks}.
\newblock \bibinfo{journal}{\emph{Social Network Analysis and Mining}}
  \bibinfo{volume}{6}, \bibinfo{number}{1} (\bibinfo{year}{2016}).
\newblock


\bibitem[\protect\citeauthoryear{Li, Ahmed, Ravi, and Smola}{Li
  et~al\mbox{.}}{2014}]%
        {li2014reducing}
\bibfield{author}{\bibinfo{person}{Aaron~Q Li}, \bibinfo{person}{Amr Ahmed},
  \bibinfo{person}{Sujith Ravi}, {and} \bibinfo{person}{Alexander~J Smola}.}
  \bibinfo{year}{2014}\natexlab{}.
\newblock \showarticletitle{Reducing the sampling complexity of topic models}.
  In \bibinfo{booktitle}{\emph{Proceedings of the 20th ACM SIGKDD international
  conference on Knowledge discovery and data mining}}. ACM,
  \bibinfo{pages}{891--900}.
\newblock


\bibitem[\protect\citeauthoryear{Maaten and Hinton}{Maaten and Hinton}{2008}]%
        {maaten2008visualizing}
\bibfield{author}{\bibinfo{person}{L Maaten} {and} \bibinfo{person}{G Hinton}.}
  \bibinfo{year}{2008}\natexlab{}.
\newblock \showarticletitle{Visualizing data using t-SNE}.
\newblock \bibinfo{journal}{\emph{Journal of machine learning research}}
  \bibinfo{volume}{9}, \bibinfo{number}{Nov} (\bibinfo{year}{2008}).
\newblock


\bibitem[\protect\citeauthoryear{Mikolov, Sutskever, Chen, Corrado, and
  Dean}{Mikolov et~al\mbox{.}}{2013}]%
        {mikolov2013distributed}
\bibfield{author}{\bibinfo{person}{Tomas Mikolov}, \bibinfo{person}{Ilya
  Sutskever}, \bibinfo{person}{Kai Chen}, \bibinfo{person}{Greg~S Corrado},
  {and} \bibinfo{person}{Jeff Dean}.} \bibinfo{year}{2013}\natexlab{}.
\newblock \showarticletitle{Distributed representations of words and phrases
  and their compositionality}. In \bibinfo{booktitle}{\emph{Advances in neural
  information processing systems}}. \bibinfo{pages}{3111--3119}.
\newblock


\bibitem[\protect\citeauthoryear{Noulas, Scellato, Mascolo, and Pontil}{Noulas
  et~al\mbox{.}}{2011}]%
        {noulas2011empirical}
\bibfield{author}{\bibinfo{person}{Anastasios Noulas},
  \bibinfo{person}{Salvatore Scellato}, \bibinfo{person}{Cecilia Mascolo},
  {and} \bibinfo{person}{Massimiliano Pontil}.}
  \bibinfo{year}{2011}\natexlab{}.
\newblock \showarticletitle{An empirical study of geographic user activity
  patterns in foursquare.}
\newblock \bibinfo{journal}{\emph{ICwSM}} \bibinfo{volume}{11},
  \bibinfo{number}{70-573} (\bibinfo{year}{2011}), \bibinfo{pages}{2}.
\newblock


\bibitem[\protect\citeauthoryear{Perozzi, Al-Rfou, and Skiena}{Perozzi
  et~al\mbox{.}}{2014}]%
        {perozzi2014deepwalk}
\bibfield{author}{\bibinfo{person}{Bryan Perozzi}, \bibinfo{person}{Rami
  Al-Rfou}, {and} \bibinfo{person}{Steven Skiena}.}
  \bibinfo{year}{2014}\natexlab{}.
\newblock \showarticletitle{Deepwalk: Online learning of social
  representations}. In \bibinfo{booktitle}{\emph{Proceedings of the 20th ACM
  SIGKDD international conference on Knowledge discovery and data mining}}.
  ACM, \bibinfo{pages}{701--710}.
\newblock


\bibitem[\protect\citeauthoryear{Recht, Re, Wright, and Niu}{Recht
  et~al\mbox{.}}{2011}]%
        {recht2011hogwild}
\bibfield{author}{\bibinfo{person}{Benjamin Recht},
  \bibinfo{person}{Christopher Re}, \bibinfo{person}{Stephen Wright}, {and}
  \bibinfo{person}{Feng Niu}.} \bibinfo{year}{2011}\natexlab{}.
\newblock \showarticletitle{Hogwild: A lock-free approach to parallelizing
  stochastic gradient descent}. In \bibinfo{booktitle}{\emph{Advances in neural
  information processing systems}}. \bibinfo{pages}{693--701}.
\newblock


\bibitem[\protect\citeauthoryear{Sun, Yin, and Ren}{Sun et~al\mbox{.}}{2017}]%
        {sun2017recommendation}
\bibfield{author}{\bibinfo{person}{Yizhou Sun}, \bibinfo{person}{Hongzhi Yin},
  {and} \bibinfo{person}{Xiang Ren}.} \bibinfo{year}{2017}\natexlab{}.
\newblock \showarticletitle{Recommendation in context-rich environment: An
  information network analysis approach}. In
  \bibinfo{booktitle}{\emph{Proceedings of the 26th International Conference on
  World Wide Web Companion}}. International World Wide Web Conferences Steering
  Committee, \bibinfo{pages}{941--945}.
\newblock


\bibitem[\protect\citeauthoryear{Tang, Qu, and Mei}{Tang
  et~al\mbox{.}}{2015a}]%
        {tang2015pte}
\bibfield{author}{\bibinfo{person}{Jian Tang}, \bibinfo{person}{Meng Qu}, {and}
  \bibinfo{person}{Qiaozhu Mei}.} \bibinfo{year}{2015}\natexlab{a}.
\newblock \showarticletitle{Pte: Predictive text embedding through large-scale
  heterogeneous text networks}. In \bibinfo{booktitle}{\emph{Proceedings of the
  21th ACM SIGKDD International Conference on Knowledge Discovery and Data
  Mining}}. ACM, \bibinfo{pages}{1165--1174}.
\newblock


\bibitem[\protect\citeauthoryear{Tang, Qu, Wang, Zhang, Yan, and Mei}{Tang
  et~al\mbox{.}}{2015b}]%
        {tang2015line}
\bibfield{author}{\bibinfo{person}{Jian Tang}, \bibinfo{person}{Meng Qu},
  \bibinfo{person}{Mingzhe Wang}, \bibinfo{person}{Ming Zhang},
  \bibinfo{person}{Jun Yan}, {and} \bibinfo{person}{Qiaozhu Mei}.}
  \bibinfo{year}{2015}\natexlab{b}.
\newblock \showarticletitle{Line: Large-scale information network embedding}.
  In \bibinfo{booktitle}{\emph{Proceedings of the 24th International Conference
  on World Wide Web}}. International World Wide Web Conferences Steering
  Committee, \bibinfo{pages}{1067--1077}.
\newblock


\bibitem[\protect\citeauthoryear{Wang, Yin, Chen, Sun, Sadiq, and Zhou}{Wang
  et~al\mbox{.}}{2015}]%
        {wang2015geo}
\bibfield{author}{\bibinfo{person}{Weiqing Wang}, \bibinfo{person}{Hongzhi
  Yin}, \bibinfo{person}{Ling Chen}, \bibinfo{person}{Yizhou Sun},
  \bibinfo{person}{Shazia Sadiq}, {and} \bibinfo{person}{Xiaofang Zhou}.}
  \bibinfo{year}{2015}\natexlab{}.
\newblock \showarticletitle{Geo-SAGE: A geographical sparse additive generative
  model for spatial item recommendation}. In
  \bibinfo{booktitle}{\emph{Proceedings of the 21th ACM SIGKDD International
  Conference on Knowledge Discovery and Data Mining}}. ACM,
  \bibinfo{pages}{1255--1264}.
\newblock


\bibitem[\protect\citeauthoryear{Wang, Yin, Sadiq, Chen, Xie, and Zhou}{Wang
  et~al\mbox{.}}{2016}]%
        {wang2016spore}
\bibfield{author}{\bibinfo{person}{Weiqing Wang}, \bibinfo{person}{Hongzhi
  Yin}, \bibinfo{person}{Shazia Sadiq}, \bibinfo{person}{Ling Chen},
  \bibinfo{person}{Min Xie}, {and} \bibinfo{person}{Xiaofang Zhou}.}
  \bibinfo{year}{2016}\natexlab{}.
\newblock \showarticletitle{Spore: A sequential personalized spatial item
  recommender system}. In \bibinfo{booktitle}{\emph{Data Engineering (ICDE),
  2016 IEEE 32nd International Conference on}}. IEEE,
  \bibinfo{pages}{954--965}.
\newblock


\bibitem[\protect\citeauthoryear{Xie, Yin, Wang, Xu, Chen, and Wang}{Xie
  et~al\mbox{.}}{2016}]%
        {xie2016learning}
\bibfield{author}{\bibinfo{person}{Min Xie}, \bibinfo{person}{Hongzhi Yin},
  \bibinfo{person}{Hao Wang}, \bibinfo{person}{Fanjiang Xu},
  \bibinfo{person}{Weitong Chen}, {and} \bibinfo{person}{Sen Wang}.}
  \bibinfo{year}{2016}\natexlab{}.
\newblock \showarticletitle{Learning graph-based poi embedding for
  location-based recommendation}. In \bibinfo{booktitle}{\emph{Proceedings of
  the 25th ACM International on Conference on Information and Knowledge
  Management}}. ACM, \bibinfo{pages}{15--24}.
\newblock


\bibitem[\protect\citeauthoryear{Yang, Bai, Zhang, Yuan, and Han}{Yang
  et~al\mbox{.}}{2017}]%
        {yang2017bridging}
\bibfield{author}{\bibinfo{person}{Carl Yang}, \bibinfo{person}{Lanxiao Bai},
  \bibinfo{person}{Chao Zhang}, \bibinfo{person}{Quan Yuan}, {and}
  \bibinfo{person}{Jiawei Han}.} \bibinfo{year}{2017}\natexlab{}.
\newblock \showarticletitle{Bridging Collaborative Filtering and
  Semi-Supervised Learning: A Neural Approach for POI Recommendation}. In
  \bibinfo{booktitle}{\emph{Proceedings of the 23rd ACM SIGKDD International
  Conference on Knowledge Discovery and Data Mining}}. ACM,
  \bibinfo{pages}{1245--1254}.
\newblock


\bibitem[\protect\citeauthoryear{Yao}{Yao}{2018}]%
        {yao2018exploiting}
\bibfield{author}{\bibinfo{person}{Zijun Yao}.}
  \bibinfo{year}{2018}\natexlab{}.
\newblock \showarticletitle{Exploiting Human Mobility Patterns for
  Point-of-Interest Recommendation}. In \bibinfo{booktitle}{\emph{Proceedings
  of the Eleventh ACM International Conference on Web Search and Data Mining}}.
  ACM, \bibinfo{pages}{757--758}.
\newblock


\bibitem[\protect\citeauthoryear{Yin, Wang, Wang, Chen, and Zhou}{Yin
  et~al\mbox{.}}{2017}]%
        {yin2017spatial}
\bibfield{author}{\bibinfo{person}{H Yin}, \bibinfo{person}{W Wang},
  \bibinfo{person}{H Wang}, \bibinfo{person}{L Chen}, {and} \bibinfo{person}{X
  Zhou}.} \bibinfo{year}{2017}\natexlab{}.
\newblock \showarticletitle{Spatial-Aware Hierarchical Collaborative Deep
  Learning for POI Recommendation}.
\newblock \bibinfo{journal}{\emph{IEEE Transactions on Knowledge and Data
  Engineering}} \bibinfo{volume}{29}, \bibinfo{number}{11}
  (\bibinfo{year}{2017}).
\newblock


\bibitem[\protect\citeauthoryear{Zhang, Zhang, Yuan, Tao, Zhang, Hanratty, and
  Han}{Zhang et~al\mbox{.}}{2017}]%
        {zhang2017react}
\bibfield{author}{\bibinfo{person}{Chao Zhang}, \bibinfo{person}{Keyang Zhang},
  \bibinfo{person}{Quan Yuan}, \bibinfo{person}{Fangbo Tao},
  \bibinfo{person}{Luming Zhang}, \bibinfo{person}{Tim Hanratty}, {and}
  \bibinfo{person}{Jiawei Han}.} \bibinfo{year}{2017}\natexlab{}.
\newblock \showarticletitle{ReAct: Online Multimodal Embedding for
  Recency-Aware Spatiotemporal Activity Modeling}. In
  \bibinfo{booktitle}{\emph{Proceedings of the 40th International ACM SIGIR
  Conference on Research and Development in Information Retrieval}}. ACM,
  \bibinfo{pages}{245--254}.
\newblock


\bibitem[\protect\citeauthoryear{Zheng, Zhang, Xie, and Ma}{Zheng
  et~al\mbox{.}}{2009}]%
        {zheng2009mining}
\bibfield{author}{\bibinfo{person}{Yu Zheng}, \bibinfo{person}{Lizhu Zhang},
  \bibinfo{person}{Xing Xie}, {and} \bibinfo{person}{Wei-Ying Ma}.}
  \bibinfo{year}{2009}\natexlab{}.
\newblock \showarticletitle{Mining interesting locations and travel sequences
  from GPS trajectories}. In \bibinfo{booktitle}{\emph{Proceedings of the 18th
  international conference on World wide web}}. ACM, \bibinfo{pages}{791--800}.
\newblock


\end{thebibliography}

\end{document}